\documentclass[journal]{IEEEtran}

\usepackage{cite}

\ifCLASSINFOpdf

\else

\fi

\usepackage{epsfig}
\usepackage{graphicx}
\usepackage{amsmath}
\usepackage{amssymb}
\usepackage{array}
\usepackage{tablefootnote}
\usepackage{subfigure}
\usepackage{float}
\usepackage{caption}
\usepackage{setspace}
\usepackage{hyperref}
\hypersetup{hypertex=true,
	colorlinks=true,
	linkcolor=blue,
	anchorcolor=blue,
	citecolor=blue}
\usepackage{algorithm}
\usepackage{algpseudocode}
\usepackage{graphics}

\usepackage{pifont}

\usepackage{multirow}

\usepackage{booktabs}
\usepackage{threeparttable}

\usepackage{amsfonts,amssymb} 

\usepackage[square,sort&compress,numbers]{natbib}

\usepackage{amsmath}

\usepackage{epstopdf}

\usepackage{makecell}

\usepackage{pifont}

\usepackage{algorithm}
\usepackage{algpseudocode}
\usepackage{soul}

\allowdisplaybreaks

\begin{document}

\title{Recursive Counterfactual Deconfounding for Object Recognition}

\author{Jiayin Sun, Hong Wang and Qiulei Dong
%
%
}


\maketitle

\begin{abstract}
Image recognition is a classic and common task in the computer vision field, which has been widely applied in the past decade. Most existing methods in literature aim to learn discriminative features from labeled images for classification, however, they generally neglect confounders that infiltrate into the learned features, resulting in low performances for discriminating test images. To address this problem, we propose a Recursive Counterfactual Deconfounding model for object recognition in both closed-set and open-set scenarios based on counterfactual analysis, called RCD. The proposed model consists of a factual graph and a counterfactual graph, where the relationships among image features, model predictions, and confounders are built and updated recursively for learning more discriminative features. It performs in a recursive manner so that subtler counterfactual features could be learned and eliminated progressively, and both the discriminability and generalization of the proposed model could be improved accordingly. In addition, a negative correlation constraint is designed for alleviating the negative effects of the counterfactual features further at the model training stage. Extensive experimental results on both closed-set recognition task and open-set recognition task demonstrate that the proposed RCD model performs better than 11 state-of-the-art baselines significantly in most cases.
\end{abstract}


\section{Introduction}    \label{section:introduction}

\IEEEPARstart{O}{bject} recognition is a basic topic in the computer vision field, and multiple deep neural networks (DNNs) based object recognition methods have achieved great success in the traditional closed-set scenario where the test label set is a subset of the training label set. Recently, object recognition deployed in an open-set scenario where samples belonging to new classes would exist in the test set has attracted more and more attention, and a lot of the corresponding methods have been proposed \cite{C2AE, GDFR, RPL, CPN, PMAL, OpenHybrid, CGDL, GMVAE-OSR, MoEP-AE, OSR_transformer2, OSR2022_3, GCPL, ARPL, OSRCI, OpenGAN, OSR2022_1, OSR2022_2, IEEE-Access, good-closed-set}, which can not only handle the closed-set recognition task but also handle the open-set recognition task.

In general, most of these methods aim to boost the feature discriminability learned from the training images, which is usually achieved by pulling the features from the same class to be more compact while pushing the features from different classes to be more far from each other. However, these methods generally neglect confounders that inevitably have infiltrated into the learned features and weaken their discriminability.



In fact, such a confounder problem in the learned features has been investigated in two feature learning based methods in other similar tasks \cite{causal_analysis_method1, causal_analysis_method2}. In \cite{causal_analysis_method1}, a counterfactual analysis method was proposed for human trajectory prediction, where the environment bias was alleviated by calculating the difference between the factual and counterfactual trajectory clues. In \cite{causal_analysis_method2}, a counterfactual debiasing method was proposed for multi-label recognition, where the contextual bias was alleviated by subtracting the bias features which were calculated by both prior context confounder and probability from the extracted image features. However, both the methods utilized manually-set confounders and conducted deconfounding only once. Hence, it still remains the following open problem: ``How to alleviate confounders that are subtle and flexibly varying with images in both closed-set recognition and open-set recognition?"  



To address this problem, we propose a Recursive Counterfactual Deconfounding model for object recognition, called RCD, which learns the counterfactual features from input images and alleviates their negative influence in a recursive manner. The proposed RCD model consists of two graphs: (i) a factual graph that represents the relationships among image features, model predictions, and confounders; (ii) a counterfactual graph based on the factual graph, where the relationships among the three nodes are updated recursively based on counterfactual analysis. RCD could learn subtler counterfactual features and eliminate the corresponding confounders progressively due to the recursive manner. Additionally, we design a negative correlation constraint for training the RCD model, which constrains the causal features to be negatively correlated to the counterfactual features, such that the confounders could be further alleviated.

The main contributions of this paper are summarized as:

\begin{enumerate}
		
	\item[-] We propose the Recursive Counterfactual Deconfounding model, RCD, where the learnable strategy is designed for dynamically obtaining the counterfactual features, and the recursive manner is used for progressively learning subtler counterfactual features. 
	
	\item[-] We design the negative correlation constraint between the discriminative features and the counterfactual features for further alleviating the negative effects of the counterfactual features.
	
	\item[-] We conduct extensive experiments in both closed-set and open-set scenarios for evaluating the effectiveness of the proposed RCD model. Results in Sec. \ref{Experiments} demonstrate that the proposed method can  significantly improve different backbones, and it also outperforms the 11 state-of-the-art baselines.

\end{enumerate}

The remainder of this paper is organized as follows. The existing state-of-the-art methods that handle both closed-set recognition and open-set recognition as well as counterfactual analysis methods in other visual tasks are reviewed in Sec.~\ref{Related_Work}. The proposed RCD model is described in detail in Sec.~\ref{Method}. Experimental results are given in Sec.~\ref{Experiments}. Finally, the paper is concluded in Sec.~\ref{Conclusion}.

\section{Related Works}    \label{Related_Work}

In this section, we firstly review the related works on the methods that handle both closed-set recognition and open-set recognition tasks, and then review the existing counterfactual analysis methods in other visual tasks.

\subsection{Methods Handling Both Closed-set and Open-set Recognition}


Recently, more and more image recognition methods attempt to improve the model generalization ability not only in closed-set scenario but also in open-set scenario. Zhang \emph{et al.} \cite{OpenHybrid} simultaneously trained a resflow network for calculating the likelihood scores of the images belonging to known classes as well as a classifier for classifying the latent features, either of which promoted the model discriminability of the other. Some methods adopted adversarial training for boosting the model discriminability. Inspired by background-class regularization \cite{background_class_ori}, Kong and Ramanan \cite{OpenGAN} took a discriminator for recognition, which was from a generative adversarial network trained by adding new classes in outlier datasets as the background classes. Chen \emph{et al.} \cite{ARPL} proposed to enlarge the gap between known-class features and unknown-class features by adversarial training with the reciprocal points. These points were prototypes in a one-vs-rest feature space corresponding to each known class. Some methods modeled the distributions of the known-class samples before learning discriminative models, such that a testing sample could be classified as one of the known classes if its feature distribution belonged to one of the modeled distributions, or could be rejected as unknown classes otherwise. Yang \emph{et al.} \cite{GCPL} modeled known-class feature distributions as multiple Gaussian mixture distributions by prototype learning. Cao \emph{et al.} \cite{GMVAE-OSR} modeled such distributions by Gaussian mixtures likewise, but they utilized an autoencoder called GMVAE \cite{GMVAE} for reconstructing images and constraining the feature distributions to approximate Gaussian mixtures. Sun \emph{et al.} \cite{MoEP-AE} proposed to model these distributions as multiple exponential power distributions for representing more complex distributions which could not be expressed by Gaussian mixtures. Besides, Azizmalayeri and Rohban \cite{OSR_transformer2} investigated different data augmentations and selected a combination of them that could most promote the model discriminability. Dai \emph{et al.} \cite{IEEE-Access} proposed a new classification score for recognition, called class activation mapping score, which could learn subtler information for boosting the model discriminability. Vaze \emph{et al.} \cite{good-closed-set} integrated multiple training strategies for improving the model discriminability.

Most of these methods aim to improve the feature discriminability by imposing discriminative losses, adding outlier data, modeling feature distributions, etc. However, they neglect the confounders that infiltrate into the learned image features. Unlike them, the proposed RCD model focuses on how to alleviate the negative effects of such confounders effectively via counterfactual analysis.

\subsection{Counterfactual Analysis Methods in Other Tasks}

The causal mechanism investigates the effect of changing a cause for learning causalities between this cause and the corresponding result \cite{causal1, causal2}, and it has been widely studied in the medical field and some social science fields such as economics and psychology. The counterfactual analysis is a convenient tool for applying the causal mechanism in a model, it replaces the original factual clues with a non-existing observation, and then analyzes the effect according to specific tasks \cite{counterfactual_analysis1, counterfactual_analysis2, counterfactual_analysis3}. 

Recently, counterfactual analysis has drawn more and more attention in computer vision, and some DNN-based methods which introduce the causal mechanism into their specific visual tasks for learning more robust models \cite{causal_analysis_method1, causal_analysis_method2, causal_analysis_method4, causal_analysis_method5, causal_analysis_method6, causal_analysis_method7, causal_analysis_method11, causal_analysis_method12, causal_NC1, causal_NC2, causal_NC3, causal_NC4} have been proposed. Most of these methods \cite{causal_analysis_method1, causal_analysis_method2, causal_analysis_method4, causal_analysis_method6, causal_analysis_method7, causal_analysis_method11} use the counterfactual analysis to alleviate the influence of the confounders/biases/distractors in their tasks, while the rest methods aim at learning more effective attentions \cite{causal_analysis_method5} or generating more effective samples \cite{causal_analysis_method12}. Generally, the deconfounding methods based on the causal mechanism firstly build causal graphs which indicate the causal relationships between different nodes. Then they build the confounder sets by manually-set operations (such as mixing or replacing pixels in images, setting random noises, and selecting models trained at the early stages). Finally, they alleviate the negative effects of the confounders by calculating the differences between the factual results and the counterfactual results. 

Among the aforementioned DNN-based deconfounding methods that adopt the counterfactual analysis, the methods most similar to our proposed RCD are proposed in \cite{causal_analysis_method1} and \cite{causal_analysis_method2}. Aiming at alleviating the effects of environment interaction in the human trajectory prediction task, Chen \emph{et al.} \cite{causal_analysis_method1} built a causal graph among environment interaction, history and future trajectories, and they selected counterfactual features from uniform rectilinear motion, mean and random trajectories. Aiming at contextual debiasing in the multi-label image recognition task, Liu \emph{et al.} \cite{causal_analysis_method2} built a causal graph among object representations, context information and predictions, and they selected a model at the early training stages for obtaining the counterfactual features. 

Compared with the aforementioned methods, the proposed method is distinctive in three aspects: (i) a learnable strategy for obtaining counterfactual features from the images dynamically, (ii) a designed negative correlation constraint for further eliminating the confounders, and (iii) a recursive manner for progressively learning subtler counterfactual features as well as improving the feature discriminability and generalization.

\begin{figure*}[t]
	\begin{center}
		\setlength{\abovecaptionskip}{0.cm}
		\includegraphics[height=7.6cm,width=13.5cm]{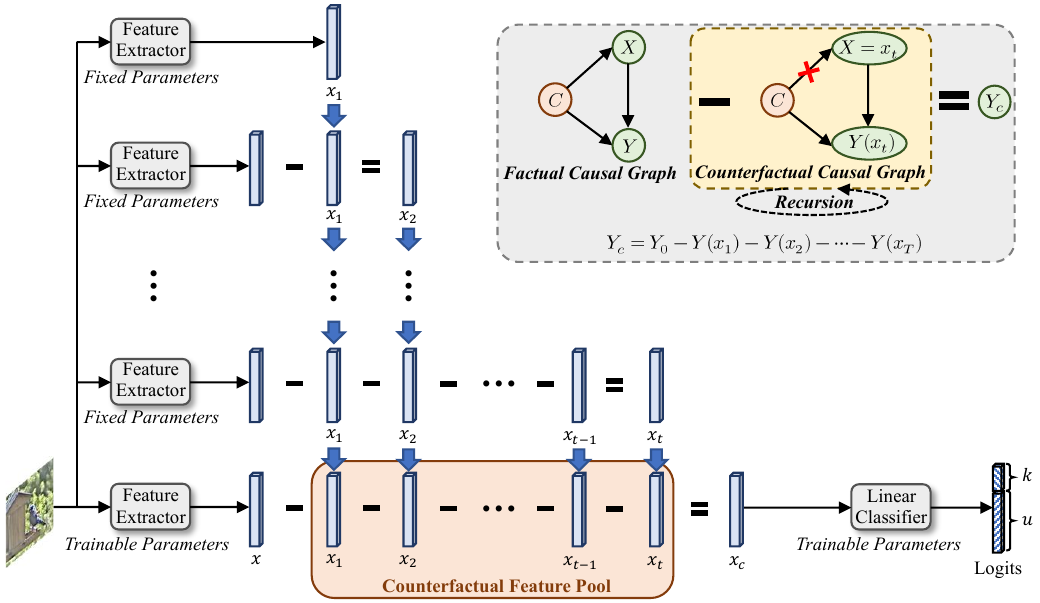}
	\end{center}
	\vspace{-8pt}
	\caption{Architecture of the proposed recursive counterfactual deconfounding model (RCD), which is used for learning the counterfactual features at the ($t+1$)-th step or for deconfounding at the $t$-th step. RCD consists of a factual graph for building the relationships among $X$ (image features), $Y$ (model predictions), and $C$ (confounders), and a counterfactual graph for updating these relationships recursively. At the $t$-th step, RCD firstly learns the counterfactual feature $x_t$ from each image and adds the feature $x_t$ to the counterfactual feature pool, then the features $\{ x_1, x_2, ..., x_t \}$ in the counterfactual feature pool are subtracted from the original feature $x$ one by one for deconfounding.}
	\label{fig: network}
	\vspace{-4pt}
\end{figure*}

\subsection{RCD for Both Closed-set Recognition and Open-set Recognition}   \label{Method}

In this section, we propose the RCD model that recursively learns and eliminates the counterfactual features. We firstly show the pipeline of the proposed method. Then, we introduce the RCD model in detail. Finally we show the training and inference strategy of the RCD model.

\subsection{Pipeline}

\begin{algorithm}[t]
	\caption{The proposed method} 
	\small
	\renewcommand{\algorithmicrequire}{\textbf{Input:}}
	\renewcommand{\algorithmicensure}{\textbf{Output:}}
	\begin{algorithmic}[1] 
		\Require The labeled training set $ \mathbf{D}^l $
		\Ensure 
		The predictions $P_T^u$ on the testing dataset $ \mathbf{D}^u $ made by the finally trained RCD model at the $T$-th step
		\State \textbf{Initialization}: Build the factual graph. Set the step number $t=1$, preset the maximum step number $T$, and prepare an empty counterfactual feature pool;
		\For{$t=1$ to $T$} 
		\State
		\textit{Learning Counterfactual Features}: Obtain the counterfactual features by training the RCD model with the loss $\mathcal{L}_{S_1}$. Add these counterfactual features to the counterfactual feature pool; 
		\State 
		\textit{Deconfounding}: Get the counterfactual features from the counterfactual feature pool, and update the counterfactual graph. Train the RCD model based on the updated counterfactual graph with the loss $\mathcal{L}_{S_2}$. Obtain the model predictions $P_t^u$;
		\EndFor 
		\State \Return $P_T^u$;
	\end{algorithmic} \label{code: pipeline}
\end{algorithm}

The pipeline of the proposed method is outlined in Algorithm \ref{code: pipeline}, and the built RCD model is shown in Fig. \ref{fig: network}. Initially, we build the factual graph among image features, model predictions, and confounders. 
At the $t$-th step ($t \in \{ 1,2,..., T \}$, where $T$ is the preset maximum step number), we firstly learn the counterfactual feature from each image based on the backbone network (when $t=1$) or based on the RCD model at the ($t-1$)-th step (when $t>1$). This feature is then added to the counterfactual feature pool. Next, the original image feature is replaced with the counterfactual feature(s) in the counterfactual feature pool, and the corresponding effect(s) is/are subtracted from the original effect. The RCD model is trained based on designed losses for updating the relationships among the three nodes, and then the model predictions on the testing dataset are obtained based on the trained model. The recursive process terminates when the step number $t$ reaches $T$, and the finally trained RCD model is evaluated.

\subsection{RCD Model} \label{model}

Here, we introduce the proposed RCD model in detail. Firstly, we build a factual causal graph based on confounders (denoted as $C$), image features (denoted as $X$), and model predictions (denoted as $Y$), for analyzing the relationships among the three elements, as shown in the upper right-hand side of Fig. \ref{fig: network}. Ideally, the model predictions depend on image features that contain robust discriminative information. However, the confounders infiltrated into the features disturb the model predictions. For example, if a kind of bird is often observed with branches, the model would be prone to relying on branches for predicting this kind of bird, which may cause the real causality hidden in the features to be neglected. Considering that the confounders in features are usually non-specific and multifarious, but their consequences are concrete and similar, \emph{i.e.}, leading to classification confusion. Thus, we regard the confounders as implicit, and design a learnable strategy for obtaining the counterfactual features.

Then, we build a counterfactual causal graph for obtaining more discriminative features, which is also shown in the upper right-hand side of Fig. \ref{fig: network}. Each original feature $x$ is replaced with the counterfactual feature $x_t$ at the $t$-th step, and the corresponding counterfactual effect $Y(x_t)$ is subtracted from the original effect $Y_0$. Considering that if the counterfactual deconfounding is conducted only once, the relatively subtler confounders would still be neglected, hence, the above process is implemented recursively, thus the finally obtained effect $Y_c$ after the subtraction at the $t$-th step, also called causal effect, is formulated as:
\begin{align}
Y_c = Y_0 - Y(x_1) - Y(x_2) - ... - Y(x_t)  \label{1}
\end{align}
We use a linear classifier for linearly mapping the final feature into a ($k+u$)-dimensional logit vector, where $k$ is the number of known classes in the training set, and $u$ is the preset number of expanded dimensions which are reserved for unknown classes in the testing set. Hence, regardless of the bias of the linear classifier, Eqn. (\ref{1}) can be re-formulated as:
\begin{align}
Y_c &= Ax - Ax_1 - Ax_2 - ... - Ax_t  \nonumber \\
&= A(x-x_1-x_2-...-x_t) = A x_c  \label{2}
\end{align}
where $A$ is the weight matrix of the linear classifier, and $x_c$ is called the causal feature. As seen from Eqn. (\ref{2}), the causal effect is equivalent to the effect of the causal feature, which is obtained by subtracting the counterfactual features one by one from the original feature. Since the causal effect alleviates the effects of the counterfactual features, the causal feature removes the confusing information from the confounders.

\vspace{-0.3cm}

\subsection{Losses and Inference}

Here, we introduce the designed losses used to train the RCD model and the inference strategy for not only closed-set recognition but also open-set recognition:

\textbf{Losses and Training.} At each step, we firstly train the RCD model to be easily confused between different categories, for learning the counterfactual features. Considering that the counterfactual feature pool is empty when $t=1$, hence, we train the network with a simple loss that causes the classification confusion, the maximum entropy loss $\mathcal{L}_{ME}$, for learning the counterfactual features from input images at the first step. This loss constrains each value in the logit vector to be close to other values:
\begin{align}
\mathcal{L}_{ME} = \frac{1}{N} \sum_{i=1}^{k+u} \sum_{j=1}^{N} p_{ij} \mathrm{log} p_{ij}   \label{3}
\end{align}
where $N$ is the batch size, $p_{ij}$ is the probability of the $j$-th image to be classified as the $i$-th class, which is calculated by \textit{SoftMax} function from the logit vector.

At the $t$-th ($t>1$) step, since the causal features are calculated based on the counterfactual features, the RCD model is trained for learning new counterfactual features with three loss terms, including a causal effect loss $\mathcal{L}_{CE}$, a negative correlation loss $\mathcal{L}_{NC}$, and a potential confounder learning loss $\mathcal{L}_{PCL}$: 
\begin{enumerate}
	\item[-] $\mathcal{L}_{CE}$: The causal effect loss term is used for alleviating the effects of the old counterfactual features. As done in counterfactual analysis methods \cite{causal_analysis_method1, causal_analysis_method2}, it encourages the logit values corresponding to the ground truth classes to be the largest, \emph{i.e.}, to take an absolute advantage in all values in the corresponding logit vector:
	\begin{align}
	\mathcal{L}_{CE} = - \frac{1}{N} \sum_{i=1}^{k+u} \sum_{j=1}^{N} y_{ij} \mathrm{log} p_{ij}  \label{4}
	\end{align}
	where $y_{ij}$ is the $i$-th value in the one-hot label corresponding to the $j$-th image.
	
	\item[-] $\mathcal{L}_{NC}$: The negative correlation loss term is used for further eliminating the old counterfactual features. At the $t$-th step ($t>1$), each causal feature is obtained after ($t-1$) subtractions. Inspired by the success of negative correlation constraint in ensemble learning \cite{negative1, negative2}, aiming at enlarging the discrepancy between two features, $\mathcal{L}_{NC}$ constrains the obtained feature after each subtraction to be negatively correlated with the corresponding counterfactual feature:
	\begin{align}
	\mathcal{L}_{NC} = \frac{1}{t-1} \sum_{N_t=1}^{t-1} \mathrm{COR} ( \boldsymbol{x}_{N_t}, \boldsymbol{x} - \boldsymbol{x}_1 - ... - \boldsymbol{x}_{N_t} )  \label{5}
	\end{align}	
	where $\mathrm{COR}$ represents the Pearson correlation coefficient between the $N_t$-th counterfactual feature $\boldsymbol{x}_{N_t}$ at the $t$-th step and the corresponding feature ($\boldsymbol{x} - \boldsymbol{x}_1 - ... - \boldsymbol{x}_{N_t}$) after the $N_t$-th subtraction. 
	
	\item[-] $\mathcal{L}_{PCL}$: 
	The potential confounder learning loss term is designed for learning the potential subtler counterfactual features (\emph{i.e.}, new counterfactual features) based on the current counterfactual analysis. This loss term controls the second largest value in a logit vector to be as close as possible to the maximum value, \emph{i.e.}, it encourages the model to learn features that are easily confused between the ground-truth class and another class:
	\begin{align}
	\mathcal{L}_{PCL} = \Vert \max\limits_{i \in \mathcal{S}^k \cup \mathcal{S}^u } f_i(\boldsymbol{x}_c) - \max\limits_{i \in \mathcal{S}^k \backslash y \cup \mathcal{S}^u} f_i(\boldsymbol{x}_c)  \Vert_2  \label{6}
	\end{align}
	where $\mathcal{S}^k$ is the index set of known-class dimensions of the logit vector, \emph{i.e.}, $\mathcal{S}^k = \{ 1,2,..., k \}$, while $\mathcal{S}^u$ is the index set of expanded dimensions of the logit vector, \emph{i.e.}, $\mathcal{S}^u = \{ k+1,k+2,..., k+u \}$. $\mathcal{S}^k \backslash y$ means removing the index of the ground truth from $\mathcal{S}^k$, $f_i(\boldsymbol{x}_c)$ is the $i$-th value of the logit vector.
	
\end{enumerate}

Hence, the total loss function used to train the RCD model for learning the counterfactual features at the $t$-th step is organized as:
\begin{align}
\left\{
\begin{aligned}
\mathcal{L}_{S_1} &= \mathcal{L}_{ME}, \ \ \ \ \ \ \ \ \ \ \ \ \ \ \ \ \ \ \ \ \ \ \ \ \ \ \ \, \mathrm{if} \  t = 1 \\
\mathcal{L}_{S_1} &= \mathcal{L}_{CE} + \lambda_1 \mathcal{L}_{NC} + \lambda_2 \mathcal{L}_{PCL}, \ \  \mathrm{if} \ t > 1  \\ 
\end{aligned}   \label{7}
\right.
\end{align}
where $\lambda_1$ and $\lambda_2$ are two hyper-parameters for balancing the three loss terms.

After learning the counterfactual features, we train the RCD model for deconfounding. Specifically, the causal effect loss term $\mathcal{L}_{CE}$ is reused for maximizing the causal effect. Besides, the negative correlation loss term $\mathcal{L}_{NC}$ is slightly modified for additionally eliminating the newly learned counterfactual features:
\begin{align}
\mathcal{L}_{NC}^{\prime} = \frac{1}{t} \sum_{N_t=1}^{t} \mathrm{COR} ( \boldsymbol{x}_{N_t}, \boldsymbol{x} - \boldsymbol{x}_1 - ... - \boldsymbol{x}_{N_t} )  \label{8}
\end{align}	

Hence, the total loss function used to train the RCD model for deconfounding at the $t$-th step is organized as:
\begin{align}
\mathcal{L}_{S_2} = \mathcal{L}_{CE} + \lambda_1 \mathcal{L}_{NC}^{\prime}   \label{9}
\end{align}

\textbf{Inference}. The logit vector mapped from the causal feature is used for inference. Specifically, the maximum value in the logit vector is used as the score (considering that unknown-class information is unavailable at the training stage, we take the maximum value in the known-class dimensions):
\begin{align}
score = \max\limits_{i \in \mathcal{S}^k} \{ f_i(\boldsymbol{x}_c) \}  \label{10}
\end{align}

In the closed-set recognition task, the model would predict the image as the known class corresponding to the score index of the logit vector. In the open-set recognition task, although unknown-class information is absent in training, expanding $u$ dimensions of the logit vectors is beneficial for boosting the model discriminability, whose effectiveness is demonstrated in Sec. \ref{Hyperparameter}. In this case, the score is compared with a threshold $\theta$ for deciding whether a testing image belongs to one of the known classes or unknown classes: If the score is larger than $\theta$, the test image would be predicted as the known class corresponding to the score index in the logit vector as done in the closed-set scenario; Otherwise, this image would be predicted as an unknown-class image. The above inference strategy in the open-set scenario is formulated as:
\begin{align}
prediction=\left\{
\begin{aligned}
&\arg\max\limits_{i \in \mathcal{S}^k} \{ f_i(\boldsymbol{x}_c) \},  \ \, \; \; \ \ \ \mathrm{if} \ score \geq \theta    \\
&unknown \ classes, \ \ \ \ \ \mathrm{if} \ score < \theta   \\
\end{aligned}  \label{11}
\right.
\end{align}

\section{Experiments}  \label{Experiments}
In this section, firstly, we introduce the dataset settings and the metrics. Next, we give the implementation details. Then, we evaluate the proposed method in both the closed-set scenario and the open-set scenario. Next, we analyze the hyper-parameters in the proposed method, and conduct the ablation study. Finally, we show the visualization results.

\vspace{-0.3cm}

\subsection{Dataset and Metrics}

\subsubsection{Datasets}

\begin{table*}[t]\footnotesize
	\renewcommand\arraystretch{0.5}
	\centering
	\setlength{\abovecaptionskip}{0cm}  
	\setlength{\belowcaptionskip}{-0.2cm} 
	\caption{Comparison of closed-set recognition results (ACC) and open-set recognition results (AUROC and OSCR) under the standard-dataset setting on Aircraft.}
	\begin{tabular}{m{1.7cm}<{\centering}m{3.3cm}<{\raggedright}m{1.5cm}<{\centering}m{3.8cm}<{\centering}m{3.8cm}<{\centering}}
		\toprule
		Backbone & Method & ACC & \makecell[c]{AUROC\\ (Easy/Medium/Hard)} & \makecell[c]{OSCR\\ (Easy/Medium/Hard)} \\ 
		\midrule
		\multirow{12}{*}{CNN} & Backbone & 0.729 & 0.830/0.796/0.663 & 0.685/0.647/0.588   \\
		& OpenHybrid \cite{OpenHybrid} & 0.735 & 0.884/0.844/0.780 & 0.718/0.687/0.629   \\
		& OpenGAN \cite{OpenGAN} & 0.807 & 0.841/0.821/0.688 & 0.795/0.754/0.651  \\
		& ARPL \cite{ARPL} & 0.917 & 0.867/0.852/0.674 & 0.814/0.820/0.657 \\
		& GCPL \cite{GCPL} & 0.823 & 0.837/0.828/0.709 & 0.805/0.761/0.652 \\
		& GMVAE-OSR \cite{GMVAE-OSR} & 0.814 & 0.849/0.833/0.687 & 0.799/0.753/0.661 \\
		& CAMV \cite{IEEE-Access} & 0.868 & 0.880/0.844/0.730 & 0.801/0.772/0.680 \\
		& Cross-Entropy+ \cite{good-closed-set} & 0.917 & 0.907/0.864/0.776 & 0.868/0.831/0.754  \\
		\cmidrule{2-5}
		& Chen \emph{et al.} \cite{causal_analysis_method1} (random) & 0.883 & 0.889/0.873/0.733 & 0.825/0.814/0.705  \\
		& Chen \emph{et al.} \cite{causal_analysis_method1} (mean) & 0.886 & 0.892/0.868/0.749 & 0.833/0.825/0.721  \\
		& Liu \emph{et al.} \cite{causal_analysis_method2} & 0.890 & 0.899/0.880/0.761 & 0.839/0.837/0.724  \\
		\cmidrule{2-5}
		& RCD & \textbf{0.918} & \textbf{0.928/0.901/0.787} & \textbf{0.881/0.857/0.756}   \\
		\midrule
		\multirow{10}{*}{SwinB} & Backbone & 0.896 & 0.848/0.827/0.723 & 0.799/0.778/0.686   \\
		& OpenHybrid & 0.901 & 0.889/0.830/0.733 & 0.823/0.781/0.689  \\
		& ARPL & 0.915 & 0.880/0.833/0.720 & 0.836/0.793/0.691   \\
		& Cross-Entropy+ & 0.903 & 0.859/0.841/0.712 & 0.814/0.796/0.681  \\
		& Trans-AUG \cite{OSR_transformer2} & 0.898 & 0.873/0.842/0.739 & 0.814/0.787/0.689  \\
		& MoEP-AE-OSR \cite{MoEP-AE} & 0.894 & 0.907/0.876/0.752 & 0.831/0.805/0.701   \\
		\cmidrule{2-5}
		& Chen \emph{et al.} \cite{causal_analysis_method1} (random) & 0.903 & 0.910/0.870/0.758 & 0.840/0.811/0.727  \\
		& Chen \emph{et al.} \cite{causal_analysis_method1} (mean) & 0.907 & 0.915/0.867/0.774 & 0.851/0.822/0.743  \\
		& Liu \emph{et al.} \cite{causal_analysis_method2} & 0.910 & 0.918/0.874/0.779 & 0.869/0.836/0.745  \\
		\cmidrule{2-5}
		& RCD & \textbf{0.932} & \textbf{0.946/0.895/0.814} & \textbf{0.900/0.855/0.785}  \\
		\bottomrule
	\end{tabular}
	\label{table: Aircraft}
\end{table*}

Similar to the common settings in the methods that handle both closed-set recognition and open-set recognition \cite{MoEP-AE, OpenHybrid, ARPL}, the experiments are conducted under two dataset settings: the standard-dataset setting and the cross-dataset setting. Both the dataset settings are based on three datasets: Aircraft \cite{Aircraft}, CUB \cite{CUB}, and Stanford-Cars \cite{Stanford-Cars}. Here, we introduce the two dataset settings respectively:

\textbf{Standard-Dataset Setting.} This setting is deployed based on the three public larger-scale image datasets in the following manner, where known-class images and unknown-class images are taken from the same dataset:
\begin{enumerate}
	\item[-] \textbf{Aircraft}: FGVC-Aircraft-2013b \cite{Aircraft} is one of the benchmark datasets for recognizing aircraft images. It comprises 10200 aircraft images, the hierarchy of whose labels contains three levels: `manufacturer' (\emph{e.g.}, 'Boeing' and 'DHC'), `family' (\emph{e.g.}, `Boeing-737' and `DHC-8'), and `variant' (\emph{e.g.}, `Boeing-737-500' and `Boeing-737-600'). We use the labels at the `variant' level, where the images are originally categorized into 100 classes. Under the standard-dataset setting, the 100 classes are divided into known classes and unknown classes. As done in the state-of-the-art method \cite{good-closed-set}, the split manner of known/unknown classes is set under three difficulty modes: `Easy', `Medium', and `Hard', according to the similarity of the labels between known classes and unknown classes. We use the same data split manner as done in \cite{good-closed-set}, where 50 classes are selected as known classes and the rest classes are further split into 20/17/13 classes as unknown classes under the three modes, respectively.
	
	\item[-] \textbf{CUB}: Caltech-UCSD Birds-200-2011 (CUB) \cite{CUB} is a common benchmark for recognizing bird images. It contains 11788 bird images belonging to 200 subclasses, \emph{e.g.}, `Sooty Albatross', `Laysan Albatross', and `Parakeet Auklet', etc., each class has its attributes. Likewise, the three difficulty modes are deployed according to the similarity between the attributes of known classes and unknown classes. We follow the data split manner as done in \cite{good-closed-set}, \emph{i.e.}, where 100 classes are selected as known classes, and 32/34/34 of the rest classes are taken as unknown classes. 
	
	\item[-] \textbf{Stanford-Cars}: Stanford-Cars \cite{Stanford-Cars} contains 16185 car images, which are divided into 196 categories according to the `Make', `Model', `Year' of the cars, \emph{e.g.}, `Acura TL Sedan 2012', `Acura TL Type-S 2008', and `Acura TSX Sedan 2012', etc. We select the first 98 classes as known classes, while the rest 98 classes are taken as unknown classes.
	
\end{enumerate}

As done in \cite{C2AE, GDFR, RPL, CPN, PMAL, OpenHybrid, CGDL, GMVAE-OSR, MoEP-AE, OSR_transformer2, OSR2022_3, GCPL, ARPL, OSRCI, OpenGAN, OSR2022_1, OSR2022_2, IEEE-Access, good-closed-set}, only the known-class images exist in the test set in the closed-set scenario, while both known-class and unknown-class images exist in the test set in the open-set scenario. To better evaluate the model performance under the semantic shift, we also deploy a cross-dataset setting for open-set recognition as done in \cite{MoEP-AE, OpenHybrid, ARPL}.

\begin{table*}\footnotesize
	\renewcommand\arraystretch{0.5}
	\centering
	\setlength{\abovecaptionskip}{0cm}  
	\setlength{\belowcaptionskip}{-0.2cm} 
	\caption{Comparison of closed-set recognition results (ACC) and open-set recognition results (AUROC and OSCR) under the standard-dataset setting on CUB.}
	\begin{tabular}{m{1.7cm}<{\centering}m{3.3cm}<{\raggedright}m{1.5cm}<{\centering}m{3.8cm}<{\centering}m{3.8cm}<{\centering}}
		\toprule
		Backbone & Method & ACC & \makecell[c]{AUROC\\ (Easy/Medium/Hard)} & \makecell[c]{OSCR\\ (Easy/Medium/Hard)}  \\
		\midrule
		\multirow{12}{*}{CNN} & Backbone & 0.675 & 0.788/0.729/0.624 & 0.632/0.586/0.502   \\
		& OpenHybrid \cite{OpenHybrid} & 0.683 & 0.873/0.862/0.739 & 0.662/0.649/0.531   \\
		& OpenGAN \cite{OpenGAN} & 0.799 & 0.801/0.765/0.707 & 0.725/0.706/0.648   \\		
		& ARPL \cite{ARPL} & 0.863 & 0.814/0.772/0.703 & 0.747/0.710/0.659  \\
		& GCPL \cite{GCPL} & 0.783 & 0.805/0.732/0.645 & 0.711/0.619/0.565   \\
		& GMVAE-OSR \cite{GMVAE-OSR} & 0.725 & 0.826/0.750/0.704 & 0.694/0.628/0.553   \\
		& CAMV \cite{IEEE-Access} & 0.836 & 0.845/0.802/0.709 & 0.746/0.715/0.640   \\
		& Cross-Entropy+ \cite{good-closed-set} & 0.862 & 0.883/0.823/0.763 & 0.798/0.754/0.708   \\
		\cmidrule{2-5}
		& Chen \emph{et al.} \cite{causal_analysis_method1} (random) & 0.848 & 0.875/0.848/0.776 & 0.786/0.770/0.712  \\
		& Chen \emph{et al.} \cite{causal_analysis_method1} (mean) & 0.855 & 0.880/0.843/0.786 & 0.788/0.773/0.718  \\
		& Liu \emph{et al.} \cite{causal_analysis_method2} & 0.860 & 0.882/0.854/0.789 & 0.791/0.779/0.720  \\
		\cmidrule{2-5}
		& RCD & \textbf{0.868} & \textbf{0.893/0.876/0.808} & \textbf{0.801/0.789/0.738}  \\
		\midrule
		\multirow{10}{*}{SwinB} & Backbone & 0.949 & 0.945/0.875/0.804 & 0.908/0.848/0.781   \\
		& OpenHybrid & 0.950 & 0.953/0.881/0.808 & 0.918/0.855/0.783   \\
		& ARPL & 0.952 & 0.948/0.877/0.810 & 0.912/0.850/0.788   \\
		& Cross-Entropy+ & \textbf{0.953} & 0.950/0.879/0.815 & 0.917/0.854/0.794   \\
		& Trans-AUG \cite{OSR_transformer2} & 0.950 & 0.953/0.882/0.818 & 0.914/0.854/0.795   \\
		& MoEP-AE-OSR \cite{MoEP-AE} & 0.948 & 0.957/0.889/0.814 & 0.915/0.856/0.787  \\
		\cmidrule{2-5}
		& Chen \emph{et al.} \cite{causal_analysis_method1} (random) & 0.950 & 0.955/0.889/0.817 & 0.916/0.858/0.796  \\
		& Chen \emph{et al.} \cite{causal_analysis_method1} (mean) & 0.951 & 0.956/0.882/0.825 & 0.918/0.860/0.805  \\
		& Liu \emph{et al.} \cite{causal_analysis_method2} & 0.951 & 0.959/0.893/0.827 & 0.919/0.863/0.808  \\
		\cmidrule{2-5}
		& RCD & \textbf{0.953} & \textbf{0.965/0.910/0.845} & \textbf{0.926/0.879/0.822}  \\
		\bottomrule
	\end{tabular}
	\label{table: CUB}
	\vspace{-0.5cm}
\end{table*}

\textbf{Cross-Dataset Setting.} Under this setting, known-class images and unknown-class images are taken from two different datasets. Here, the 50 known classes split from the Aircraft dataset are taken as known classes, while the whole 200 classes or 196 classes of the CUB dataset and the Stanford-Cars dataset are taken as unknown classes, respectively.

\subsubsection{Metrics}

Under the two dataset settings, the following evaluation metrics are used:



\textbf{Standard-Dataset Setting.} Under this setting, we adopt two commonly-used metrics in OSR methods \cite{MoEP-AE, OpenHybrid, OpenGAN, ARPL, OSR_transformer2}, \emph{i.e.}, ACC and AUROC, as well as OSCR as done in \cite{good-closed-set}. Here, we introduce these metrics in detail:
\begin{enumerate}
	\item[-] \textbf{ACC}: ACC is the Top-1 accuracy, which is a commonly-used metric in the closed-set recognition task. It evaluates the classification accuracy in the closed-set scenario.
	
	\item[-] \textbf{AUROC}: Area Under the Receiver Operating Characteristic curve (AUROC) is a threshold-independent metric in binary classification. Here, it measures the open-set recognition performance, which regards all of the known classes as one class and the whole unknown classes as the other class. A larger AUROC indicates better performance in distinguishing between known classes and unknown classes.
	
	\item[-] \textbf{OSCR}: Open-Set Classification Rate (OSCR) \cite{OSCR} is likewise a threshold-independent metric that measures the open-set recognition performance. Different from AUROC, it also considers the accuracy of known-class images in the open-set recognition task, and is calculated based on the curve depicted by Correct Classification Rate (CCR, which is defined as the proportion of the known-class samples that are correctly classified and have higher classification probabilities than a threshold) versus False Positive Rate (FPR, which is defined as the proportion of unknown-class samples that are mistaken as any known class with higher probabilities than the same threshold). A larger OSCR indicates better performance in both classifying known-class images and distinguishing between known-class images and unknown-class images.
	
\end{enumerate}

\textbf{Cross-Dataset Setting.} Under this setting, we adopt the macro-F1 score as the metric as done in \cite{OpenGAN, MoEP-AE, IEEE-Access}:

\begin{enumerate}
	\item[-] \textbf{macro-F1}: The macro-F1 score is a threshold-dependent metric that measures the open-set recognition performance. In accordance with \cite{GDFR, MoEP-AE}, the threshold is set to the value under which 90\% images in the validation set can be correctly recognized as known classes. Here, it regards the open-set classification task as a ($k+1$)-way classification task (where unknown classes are taken as the ($k+1$)-th class). A larger macro-F1 score indicates better performance in classifying the ($k+1$)-class images.
	
\end{enumerate}

\vspace{-0.3cm}

\subsection{Implementation Details}

The image sizes in the Aircraft, CUB, and Stanford-Cars datasets are set to $448 \times 448$, $448 \times 448$, and $224 \times 224$, respectively. A typical CNN backbone, ResNet50 \cite{DNNs2}, and a typical fully-connected backbone, SwinB \cite{Swin} are taken as the feature extractor, respectively. We use SGD optimizer with the weight decay of $1 \times 10^{-4}$. We use the learning rate $1 \times 10^{-3}$ for training CNN networks, while $2 \times 10^{-4}$ for training transformer networks. The backbones are initialized by the pretrained models on ImageNet \cite{ImageNet}. The two hyper-parameters $\lambda_1$ and $\lambda_2$ in the total loss functions are both set to $1$. The number $u$ of expanded dimensions of the logit vector is set to $1024$. The maximum step number $T$ is set to $2$.

\subsection{Evaluation Under the Standard-Dataset Setting} \label{standard_evaluation}


\begin{table}\footnotesize
	\renewcommand\arraystretch{0.5}
	\centering
	\setlength{\abovecaptionskip}{0cm}
	\caption{Comparison of closed-set recognition results (ACC) and open-set recognition results (AUROC and OSCR) under the standard-dataset setting on Stanford-Cars.}
	\begin{tabular}{m{1.1cm}<{\centering}m{3.0cm}<{\raggedright}m{0.9cm}<{\centering}m{0.8cm}<{\centering}m{0.8cm}<{\centering}}
		\toprule
		Backbone & Method & ACC & AUROC & OSCR  \\
		\midrule
		\multirow{12}{*}{CNN} & Backbone & 0.659 & 0.678 & 0.602   \\
		& OpenHybrid \cite{OpenHybrid} & 0.675 & 0.699 & 0.637 \\
		& OpenGAN \cite{OpenGAN} & 0.704 & 0.763 & 0.682 \\	
		& ARPL \cite{ARPL} & 0.752 & 0.839 & 0.726 \\
		& GCPL \cite{GCPL} & 0.688 & 0.735 & 0.663   \\
		& GMVAE-OSR \cite{GMVAE-OSR} & 0.683 & 0.756 & 0.670   \\
		& CAMV \cite{IEEE-Access} & 0.754 & 0.832 & 0.724 \\	
		& Cross-Entropy+ \cite{good-closed-set} & 0.768 & 0.859 & 0.735 \\
		\cmidrule{2-5}
		& Chen \emph{et al.} \cite{causal_analysis_method1} (random) & 0.775 & 0.858 & 0.739  \\
		& Chen \emph{et al.} \cite{causal_analysis_method1} (mean) & 0.778 & 0.864 & 0.743  \\
		& Liu \emph{et al.} \cite{causal_analysis_method2} & 0.783 & 0.867 & 0.745  \\
		\cmidrule{2-5}
		& RCD & \textbf{0.827} & \textbf{0.888} & \textbf{0.763}  \\
		\midrule
		\multirow{10}{*}{SwinB} & Backbone & 0.856 & 0.905 & 0.800 \\
		& OpenHybrid & 0.857 & 0.910 & 0.805  \\
		& ARPL & 0.860 & 0.908 & 0.807  \\
		& Cross-Entropy+ & 0.864 & 0.906 & 0.808  \\
		& Trans-AUG \cite{OSR_transformer2} & 0.860 & 0.910 & 0.809   \\
		& MoEP-AE-OSR \cite{MoEP-AE} & 0.869 & 0.922 & 0.819  \\
		\cmidrule{2-5}
		& Chen \emph{et al.} \cite{causal_analysis_method1} (random) & 0.865 & 0.913 & 0.815  \\
		& Chen \emph{et al.} \cite{causal_analysis_method1} (mean) & 0.870 & 0.916 & 0.819  \\
		& Liu \emph{et al.} \cite{causal_analysis_method2} & 0.873 & 0.918 & 0.822  \\
		\cmidrule{2-5}
		& RCD & \textbf{0.890} & \textbf{0.930} & \textbf{0.846}  \\
		\bottomrule
	\end{tabular}
	\label{table: Stanford-Cars}
	\vspace{-0.1cm}
\end{table}

We compare the proposed method with 11 existing methods. Firstly, we evaluate 9 state-of-the-art methods that handle both closed-set and open-set recognition (\emph{i.e.}, OpenHybrid \cite{OpenHybrid}, OpenGAN \cite{OpenGAN}, ARPL \cite{ARPL}, GCPL \cite{GCPL}, GMVAE-OSR \cite{GMVAE-OSR}, CAMV \cite{IEEE-Access}, Cross-Entropy+ \cite{good-closed-set}, Trans-AUG \cite{OSR_transformer2}, and MoEP-AE-OSR \cite{MoEP-AE}). To eliminate the possible influence of different feature extractors used in the comparative methods, we also evaluate three relatively better methods (\emph{i.e.}, OpenHybrid, ARPL, and Cross-Entropy+) by replacing their original CNN feature extractors with the SwinB for further comparison. In addition, we also evaluate the 2 aforementioned counterfactual deconfounding methods \cite{causal_analysis_method1, causal_analysis_method2} by replacing our counterfactual deconfounding strategy with theirs. For comprehensive evaluation, we evaluate two different counterfactual features used in \cite{causal_analysis_method1} ((i) randomly sampling from a uniform distribution, (ii) calculating the mean of all image features belonging to each class). The closed-set recognition and open-set recognition results under the standard-dataset setting on the Aircraft, CUB, and Stanford-Cars datasets are respectively reported in Tables \ref{table: Aircraft}, \ref{table: CUB}, and \ref{table: Stanford-Cars}. Besides, The testing images per second (FPS) at the inference stage on Aircraft are reported in Table \ref{table: computation}.

\begin{table}\footnotesize
	\renewcommand\arraystretch{0.5}
	\centering
	\setlength{\abovecaptionskip}{0cm}  
	\caption{Computational resources on Aircraft.}
	\begin{tabular}{m{1.8cm}<{\centering}m{3.3cm}<{\raggedright}m{1.7cm}<{\centering}}
		\toprule
		Backbone & Method & FPS  \\
		\midrule
		\multirow{12}{*}{CNN} & Backbone & 48    \\
		& OpenHybrid \cite{OpenHybrid} & 24 \\
		& OpenGAN \cite{OpenGAN} & 35 \\	
		& ARPL \cite{ARPL} & 41 \\
		& GCPL \cite{GCPL} & 42   \\
		& GMVAE-OSR \cite{GMVAE-OSR} & 39   \\
		& CAMV \cite{IEEE-Access} & 47 \\	
		& Cross-Entropy+ \cite{good-closed-set} & 46 \\
		\cmidrule{2-3}
		& Chen \emph{et al.} \cite{causal_analysis_method1} (random) & 42   \\
		& Chen \emph{et al.} \cite{causal_analysis_method1} (mean) & 42  \\
		& Liu \emph{et al.} \cite{causal_analysis_method2} & 40  \\
		\cmidrule{2-3}
		& RCD & 42  \\
		\midrule
		\multirow{10}{*}{SwinB} & Backbone & 29  \\
		& OpenHybrid & 13  \\
		& ARPL & 24  \\
		& Cross-Entropy+ & 28  \\
		& Trans-AUG \cite{OSR_transformer2} & 27   \\
		& MoEP-AE-OSR \cite{MoEP-AE} & 23  \\
		\cmidrule{2-3}
		& Chen \emph{et al.} \cite{causal_analysis_method1} (random) & 25   \\
		& Chen \emph{et al.} \cite{causal_analysis_method1} (mean) & 25  \\
		& Liu \emph{et al.} \cite{causal_analysis_method2} & 23  \\
		\cmidrule{2-3}
		& RCD & 25 \\
		\bottomrule
	\end{tabular}
	\label{table: computation}
	\vspace{-0.2cm}
\end{table}

\begin{table}[t]\footnotesize
	\renewcommand\arraystretch{0.5}
	\centering
	\setlength{\abovecaptionskip}{0cm} 
	\caption{Comparison of open-set recognition results (macro-F1) under the cross-dataset setting. Under this setting, Aircraft is used as the known-class dataset, while CUB and Stanford-Cars are used as the unknown-class datasets respectively.}
	\begin{tabular}{m{1.0cm}<{\centering}m{3.0cm}<{\raggedright}m{1cm}<{\centering}m{1.8cm}<{\centering}}
		\toprule
		Backbone & Method & CUB & Stanford-Cars \\
		\midrule
		\multirow{12}{*}{CNN} & Backbone & 0.418 & 0.795  \\
		& OpenHybrid \cite{OpenHybrid} & 0.473 & 0.840 \\
		& OpenGAN \cite{OpenGAN} & 0.426 & 0.819 \\	
		& ARPL \cite{ARPL} & 0.469 & 0.831 \\
		& GCPL \cite{GCPL} & 0.435 & 0.812  \\
		& GMVAE-OSR \cite{GMVAE-OSR} & 0.454 & 0.826   \\
		& CAMV \cite{IEEE-Access} & 0.461 & 0.833 \\	
		& Cross-Entropy+ \cite{good-closed-set} & 0.482 & 0.869 \\
		\cmidrule{2-4}
		& Chen \emph{et al.} \cite{causal_analysis_method1} (random) & 0.497 & 0.873   \\
		& Chen \emph{et al.} \cite{causal_analysis_method1} (mean) & 0.506 & 0.875  \\
		& Liu \emph{et al.} \cite{causal_analysis_method2} & 0.513 & 0.876  \\
		\cmidrule{2-4}
		& RCD & \textbf{0.551} & \textbf{0.882}  \\
		\midrule
		\multirow{10}{*}{SwinB} & Backbone & 0.497 & 0.886  \\
		& OpenHybrid & 0.499 & 0.875  \\
		& ARPL & 0.521 & 0.888  \\
		& Cross-Entropy+ & 0.503 & 0.891  \\
		& Trans-AUG \cite{OSR_transformer2} & 0.508 & 0.882 \\
		& MoEP-AE-OSR \cite{MoEP-AE} & 0.529 & 0.876 \\
		\cmidrule{2-4}
		& Chen \emph{et al.} \cite{causal_analysis_method1} (random) & 0.528 & 0.883   \\
		& Chen \emph{et al.} \cite{causal_analysis_method1} (mean) & 0.530 & 0.886  \\
		& Liu \emph{et al.} \cite{causal_analysis_method2} & 0.531 & 0.893  \\
		\cmidrule{2-4}
		& RCD & \textbf{0.568} & \textbf{0.911}   \\
		\bottomrule
	\end{tabular}
	\label{table: cross-dataset}
	\vspace{-0.5cm}
\end{table}

\begin{figure*}[t]
	\begin{center}
		\setlength{\abovecaptionskip}{0.cm}
		\includegraphics[height=4.5cm,width=17cm]{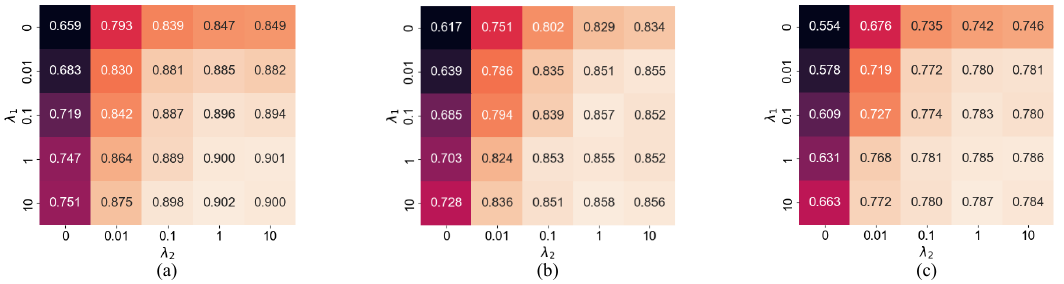}
	\end{center}
\vspace{-10pt}
	\caption{Comparison of the proposed method with different configurations of $\lambda_1 = \{ 0, 0.01, 0.1, 1, 10 \}$ and $\lambda_2 = \{ 0, 0.01, 0.1, 1, 10 \}$. The OSCR results are reported under the three difficulty modes: (a) the `Easy' mode, (b) the `Medium' mode, and (c) the `Hard' mode. The experiments are conducted on Aircraft under the standard-dataset setting, and the SwinB is used as the feature extractor. A lighter color indicates a better result.}
	\label{fig: loss_weights}
\end{figure*}

As seen from Tables \ref{table: Aircraft}-\ref{table: Stanford-Cars}, A stronger backbone helps boost the model performance on most datasets. Furthermore, with the same backbone, the proposed method achieves better performance than the comparative methods (including the methods that handle closed-set and open-set recognition as well as the two aforementioned counterfactual deconfounding methods) on all datasets in closed-set and open-set scenarios, demonstrating both the improvement of the model discriminability and the effectiveness of the proposed counterfactual deconfounding strategy. Besides, as seen from Table \ref{table: computation}, the testing efficiency of RCD is close to that of nine existing methods (ARPL, GCPL, GMVAE-OSR, CAMV, Cross-Entropy+, Trans-AUG, MoEP-AE-OSR, Chen \emph{et al.}, and Liu \emph{et al.}), but significantly higher than that of two existing methods (OpenHybrid and OpenGAN), indicating that our proposed method is competitive on testing efficiency.

%
%

\subsection{Evaluation Under the Cross-Dataset Setting} \label{cross_evaluation}

The proposed method is also evaluated under the cross-dataset setting where known-class images and unknown-class images are from different datasets in the open-set scenario. Specifically, the model is trained with known-class training images on Aircraft. Then, it is tested with both known-class testing images on Aircraft and unknown-class testing images from CUB and Stanford-Cars respectively. The results are reported in Table \ref{table: cross-dataset}. As seen from this table, RCD significantly outperforms other models, demonstrating the improvement of the model generalization.


\subsection{Analysis of the Hyper-Parameters} \label{Hyperparameter}

In this subsection, we conduct experiments for analyzing the influence of the hyper-parameters in the proposed method, including the two weighting coefficients $\lambda_1$ and $\lambda_2$ in the loss functions (\emph{i.e.}, Eqns. (\ref{7}) and (\ref{9})), the number $u$ of expanded dimensions of logit vectors, and the maximum step number $T$. The experiments are conducted on the Aircraft dataset under the standard-dataset setting, and the SwinB is used as the feature extractor.

\begin{table*}\footnotesize
	\renewcommand\arraystretch{0.5}
	\centering
	\setlength{\abovecaptionskip}{0cm} 
	\caption{Closed-set recognition results (ACC) and open-set recognition results (AUROC and OSCR) of the proposed method with different configurations of $u = \{ 0, 64, 256, 1024, 2048 \}$. The experiments are conducted on Aircraft under the standard-dataset setting, and the SwinB is used as the feature extractor.}
	\begin{tabular}{m{3cm}<{\centering}m{2.5cm}<{\centering}m{4.5cm}<{\centering}m{4.5cm}<{\centering}}
		\toprule
		$u$ & ACC & \makecell[c]{AUROC\\ (Easy/Medium/Hard)} & \makecell[c]{OSCR\\ (Easy/Medium/Hard)}  \\
		\midrule
		0 & 0.924 & 0.934/0.878/0.783 & 0.881/0.833/0.749  \\
		64 & 0.925 & 0.935/0.881/0.796 & 0.882/0.839/0.764  \\
		256 & 0.930 & 0.941/0.893/0.809 & 0.892/0.852/0.778  \\
		1024 & 0.932 & 0.946/0.895/0.814 & 0.900/0.855/0.785  \\
		2048 & 0.930 & 0.951/0.898/0.812 & 0.902/0.853/0.781  \\
		\bottomrule
	\end{tabular}
	\label{table: extend_u}
\end{table*}

\begin{table*}\footnotesize
	\renewcommand\arraystretch{0.5}
	\centering
	\setlength{\abovecaptionskip}{0cm} 
	\caption{Closed-set recognition results (ACC) and open-set recognition results (AUROC and OSCR) of the proposed method with different configurations of $T = \{ 1,2,3,4 \}$. The experiments are conducted on Aircraft under the standard-dataset setting, and the SwinB is used as the feature extractor.}
	\begin{tabular}{m{3cm}<{\centering}m{2.5cm}<{\centering}m{4.5cm}<{\centering}m{4.5cm}<{\centering}}
		\toprule
		$T$ & ACC & \makecell[c]{AUROC\\ (Easy/Medium/Hard)} & \makecell[c]{OSCR\\ (Easy/Medium/Hard)}  \\
		\midrule
		1 & 0.910 & 0.942/0.889/0.770 & 0.888/0.844/0.736  \\
		2 & 0.932 & 0.946/0.895/0.814 & 0.900/0.855/0.785  \\
		3 & 0.931 & 0.953/0.895/0.812 & 0.908/0.852/0.783  \\
		4 & 0.932 & 0.948/0.896/0.817 & 0.901/0.854/0.788  \\
		\bottomrule
	\end{tabular}
	\label{table: max_T}
	\vspace{-0.1cm}
\end{table*}

\subsubsection{Analysis of $\lambda_1$ and $\lambda_2$} \label{loss_ablation}

Here, we analyze the influence of $\lambda_1$ and $\lambda_2$ in loss functions that balance the weights of the negative correlation loss term $\mathcal{L}_{NC}$ (or $\mathcal{L}_{NC}^{\prime}$) and the potential confounder learning loss term $\mathcal{L}_{PCL}$. Specifically, we train the RCD model with $\lambda_1 = \{ 0, 0.01, 0.1, 1, 10 \}$ and $\lambda_2 = \{ 0, 0.01, 0.1, 1, 10 \}$, and the corresponding OSCR (which is a comprehensive metric that considers both the classification accuracy of known-class samples and the distinction between known-class samples and unknown-class samples) results on the testing dataset under the three difficulty modes are shown in Fig. \ref{fig: loss_weights}. Three points can be seen from this figure: 
\begin{enumerate}
	\item[-] The model performance would decrease if the potential confounder learning loss term $\mathcal{L}_{PCL}$ is absent in Eqn. (\ref{7}) when learning counterfactual features. This is because $\mathcal{L}_{PCL}$ is used for making use of subtler confounders, the counterfactual analysis couldn't work or even damages the model discriminability if the confounders are not made use of (\emph{i.e.}, $\mathcal{L}_{PCL}$ is removed).
	
	\item[-] With the increasement of $\lambda_1$, the OSCR score increases. This trend demonstrates that the negative correlation loss term $\mathcal{L}_{NC}$ (or $\mathcal{L}_{NC}^{\prime}$) helps boost the feature discriminability, since it imposes stronger constraints on eliminating the confounders (represented as the counterfactual features).
	
	\item[-] The OSCR results vary slightly when $\lambda_1$ varies in $[0.1, 10]$ and $\lambda_2$ varies in $[0.1, 10]$, indicating that the proposed method is not quite sensitive to the two hyper-parameters within these intervals.

\end{enumerate}


\begin{figure*}[t]
	\centering
	\subfigure[]{
		\centering
		\includegraphics[width=5.2cm]{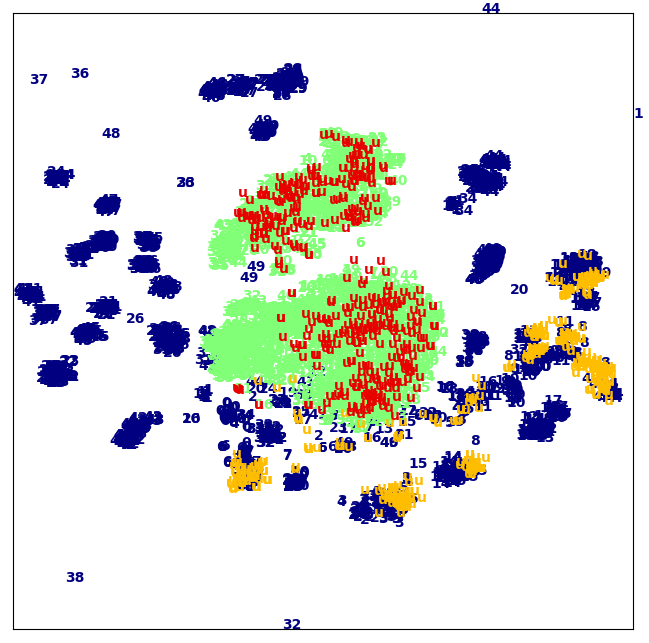}
	}
	\centering
	\subfigure[]{
		\centering
		\includegraphics[width=5.2cm]{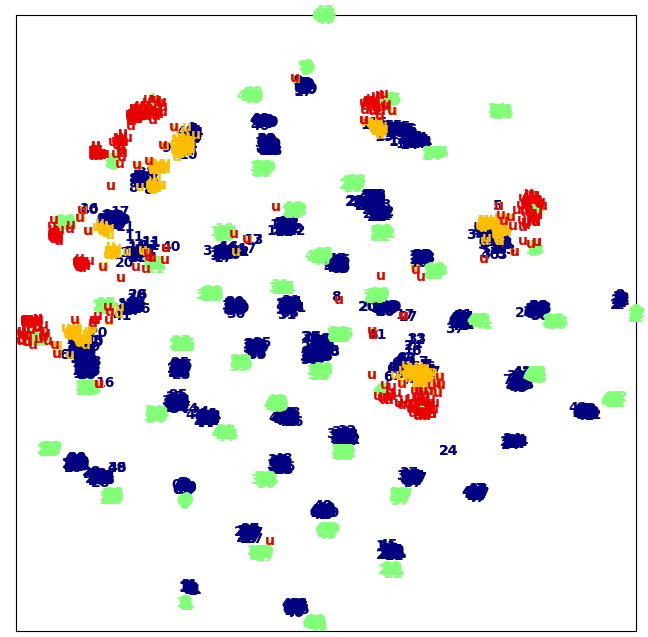}
	}
	\centering
	\subfigure[]{
		\centering
		\includegraphics[width=5.2cm]{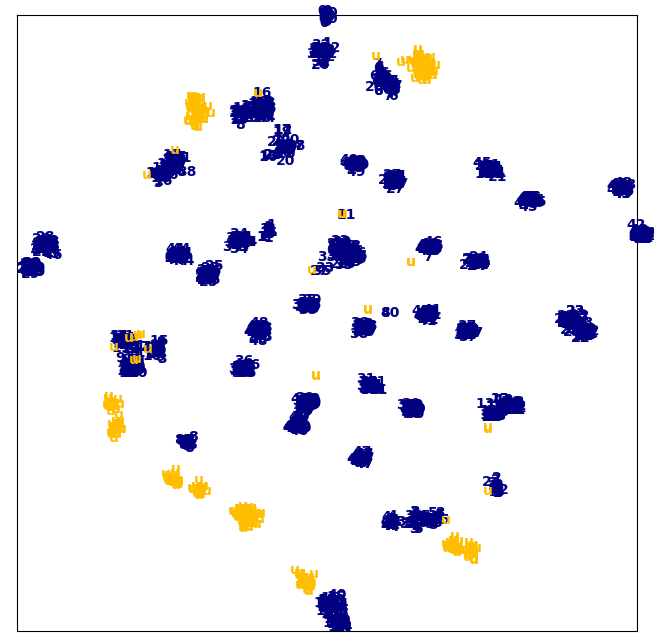}
	}
	\vspace{-6pt}
	\caption{T-SNE visualization of the learned features based on three methods: (a) backbone method, (b) the proposed method with $t=1$, and (c) the proposed method with $t=2$. The points in blue, green, yellow, and red represent known-class causal features, known-class counterfactual features, unknown-class causal features, and unknown-class counterfactual features, respectively.}
	\label{fig: visualization1}
	\vspace{-8pt}
\end{figure*}

\subsubsection{Analysis of the number $u$ of expanded dimensions of logit vectors}

It is noted that we expand logit vectors to reserve some logit dimensions for unknown classes in the open-set scenario. Here, we analyze the influence of $u$, which represents the number of expanded dimensions of logit vectors. Specifically, we evaluate the proposed method under five different configurations of $u$: $u \in \{ 0, 64, 256, 1024, 2048 \}$, whose closed-set recognition and open-set recognition results are reported in Table \ref{table: extend_u}. Two points can be seen from this table:
\begin{enumerate}
	\item[-] As $u$ increases, the open-set recognition results of RCD increase, demonstrating that expanding the dimensions of logit vectors helps handle the object recognition task in the open-set scenario, though unknown-class data is not used in training. 
	
	\item[-] Both closed-set recognition results and open-set recognition results vary slightly when $u$ varies within $[256, 2048]$, demonstrating that RCD is not sensitive to $u$ when $u$ lies within this range.
	
\end{enumerate}

\subsubsection{Analysis of the maximum step number $T$}

The proposed method is conducted recursively. Here, we analyze the influence of the maximum step number $T$. Specifically, we evaluate RCD with $T=\{ 1, 2, 3, 4 \}$, and the corresponding closed-set recognition and open-set recognition results are reported in Table \ref{table: max_T}. Two points can be seen from this table: 
\begin{enumerate}
	\item[-] The model with $T=2$ performs better than the model with $T=1$, especially on ACC and AUROC/OSCR under harder modes. Such observation indicates that the recursion is effective for boosting the model discriminability, especially in the closed-set scenario and in a more difficult open-set scenario, which also indicates that RCD could learn subtler counterfactual features and alleviate their negative effects through steps.
	
	\item[-] The model already has relatively better results when the maximum step number $T$ reaches $2$, and continuing to increase $T$ couldn't further boost the model performance in most cases, demonstrating that RCD could achieve excellent performance without relying on multiple steps.
	
\end{enumerate}

\vspace{-0.3cm}

\subsection{Visualization} \label{visualization}

In Fig. \ref{fig: visualization1}, we use t-SNE \cite{tSNE} to visualize the causal features learned by (a) the backbone method (where the extracted features are directly fed into the linear classifier, the logit vector is a $k$-dimensional vector, and the model is trained with the cross-entropy classification loss), (b) the proposed method with $t=1$, and (c) the proposed method with $t=2$. The known-class causal features and unknown-class causal features are in blue and yellow, respectively. Besides, considering that the counterfactual features at the $t$-th step are learned based on the causal features at the ($t-1$)-th step when $t>1$, hence, the counterfactual features learned when $t=1$ are visualized in Fig. \ref{fig: visualization1} (a), and the counterfactual features learned when $t=2$ are visualized in Fig. \ref{fig: visualization1} (b). The known-class counterfactual features and unknown-class counterfactual features are in green and red, respectively.

Two points can be seen from the three subfigures:
\begin{enumerate}
	\item[-] Compared with the backbone method, the proposed method at the first step learns more discriminative features, and a distinct gap between known-class causal features and unknown-class causal features can be observed, demonstrating the effectiveness of counterfactual deconfounding. Besides, the proposed method with $t=2$ further improves the discriminability of the proposed method with $t=1$, demonstrating the effectiveness of the recursive deconfounding manner.
	
	\item[-] The counterfactual features learned when $t=1$ are mixed and far away from the causal features learned by the backbone method, while the counterfactual features learned when $t=2$ are discriminative between different categories and closer to the causal features at the previous step, demonstrating that the RCD model has learned subtler counterfactual features when $t=2$.
	
\end{enumerate}

\section{Conclusion}   \label{Conclusion}

In this paper, we propose the recursive counterfactual deconfounding model for handling the object recognition task in both the closed-set scenario and the open-set scenario, called RCD, for learning more discriminative features. A learnable strategy to obtain the counterfactual features and a recursive deconfounding manner are designed for learning subtler counterfactual features (representing the confounders) and alleviating the corresponding negative effects progressively. Besides, A negative correlation constraint is designed to train the RCD model for eliminating the confounders further. Extensive experimental results demonstrate the effectiveness of the proposed method.
 
In the future, the proposed RCD model would be used in other visual tasks, such as zero-shot learning, domain adaptation, domain generalization, etc.

\ifCLASSOPTIONcaptionsoff
  \newpage
\fi

\small
\bibliographystyle{IEEEtran}
\bibliography{egbib_20230219}

%
%

\end{document}